%% file: main.tex
\documentclass[10pt,twocolumn,letterpaper]{article}

\usepackage[pagenumbers]{cvpr} 
\usepackage{indentfirst}

\usepackage{colortbl}
\usepackage{xcolor}
\usepackage{graphicx}

\usepackage{colortbl}
\usepackage{multirow}
\usepackage{booktabs}

\definecolor{cvprblue}{rgb}{0.21,0.49,0.74}
\usepackage[pagebackref,breaklinks,colorlinks,allcolors=cvprblue]{hyperref}

\def\paperID{8732} 

\def\confName{CVPR}
\def\confYear{2025}

\title{LTCF-Net: A Transformer-Enhanced Dual-Channel Fourier Framework for Low-Light Image Restoration}

\author{
Gaojing Zhang$^{1}$\\
$^{1}$University of Sussex\\
{\tt\small gz69@sussex.ac.uk}
\and
Jinglun Feng$^{2}$\thanks{Corresponding author.}\\
$^{2}$City College of New York\\
{\tt\small jfeng1@ccny.cuny.edu}
}

\begin{document}
\maketitle
\input{sec/0_abstract}    
\input{sec/1_intro}

\input{sec/2_related_work}
\input{sec/3_Method}

\input{sec/4_Experiment}
\input{sec/5_Conclusion}

{
    \small
    \newpage
    \bibliographystyle{ieeenat_fullname}
    \bibliography{main}
}

\end{document}

%% file: sec/0_abstract.tex
\begin{abstract}
We introduce LTCF-Net, a novel network architecture designed for enhancing low-light images. Unlike Retinex-based methods, our approach utilizes two color spaces—LAB and YUV—to efficiently separate and process color information, by leveraging the separation of luminance from chromatic components in color images. In addition, our model incorporates the Transformer architecture to comprehensively understand image content while maintaining computational efficiency. To dynamically balance the brightness in output images, we also introduce a Fourier transform module that adjusts the luminance channel in the frequency domain. This mechanism could uniformly balance brightness across different regions while eliminating background noises, and thereby enhancing visual quality. By combining these innovative components, LTCF-Net effectively improves low-light image quality while keeping the model lightweight. Experimental results demonstrate that our method outperforms current state-of-the-art approaches across multiple evaluation metrics and datasets, achieving more natural color restoration and a balanced brightness distribution.
\end{abstract}

%% file: sec/1_intro.tex
\section{Introduction}
Low-Light Image Enhancement (LLIE) is a critical and complex task within the domain of computer vision. In environments with inadequate lighting, camera images often exhibit severe noise, diminished contrast, and obscured details. These degraded images not only compromise human visual perception but also challenge the downstream visual tasks such as object detection at night. The primary objective of LLIE is to enhance the visibility and contrast of these images, reveal details obscured in shadows, and mitigate distortions that typically occur during the enhancement process, including noise, unwanted artifacts, and inaccurate color reproduction.

Traditional methods like histogram equalization\cite{25,20,21,23} and gamma correction\cite{22} are among the most straightforward techniques to enhance visibility in low-light images. However, while these methods effectively increase contrast, they often lead to oversaturation in bright areas but also result in a loss of detail \cite{1}, without addressing the underlying issues of inadequate illumination. In contrast, Retinex theory, which simulates human visual perception of color, has become a foundational principle in LLIE strategies \cite{8,36,37,38,39,40}. Although Retinex-based methods leverage illumination estimation and reflectance for enhancement, their underlying assumption of clean, distortion-free inputs rarely holds in real low-light conditions~\cite{36,37}. This fundamental limitation often results in amplified noise and color artifacts, compromising their practical effectiveness.

Recent learning-based approaches have attempted direct mapping between low-light and normal-light conditions~\cite{32,33,34,35}. While these methods show promise, they often sacrifice perceptual color accuracy and theoretical grounding for performance~\cite{3}, complicated by their multi-stage training requirements. The introduction of Transformers \cite{11}, known for capturing global dependencies, initially seemed promising. However, the computational demands of standard Vision Transformers~\cite{63,64} proved prohibitive~\cite{7}, leading to hybrid CNN-Transformer architectures like SNR-Net \cite{6} and LYT-Net \cite{60}. While these models incorporate global Transformer layers at reduced resolutions, they have yet to fully capitalize on Transformers' potential for low-light enhancement, indicating room for further advancement.

\begin{figure*}[ht]
    \centering
    \includegraphics[width=0.98\textwidth]{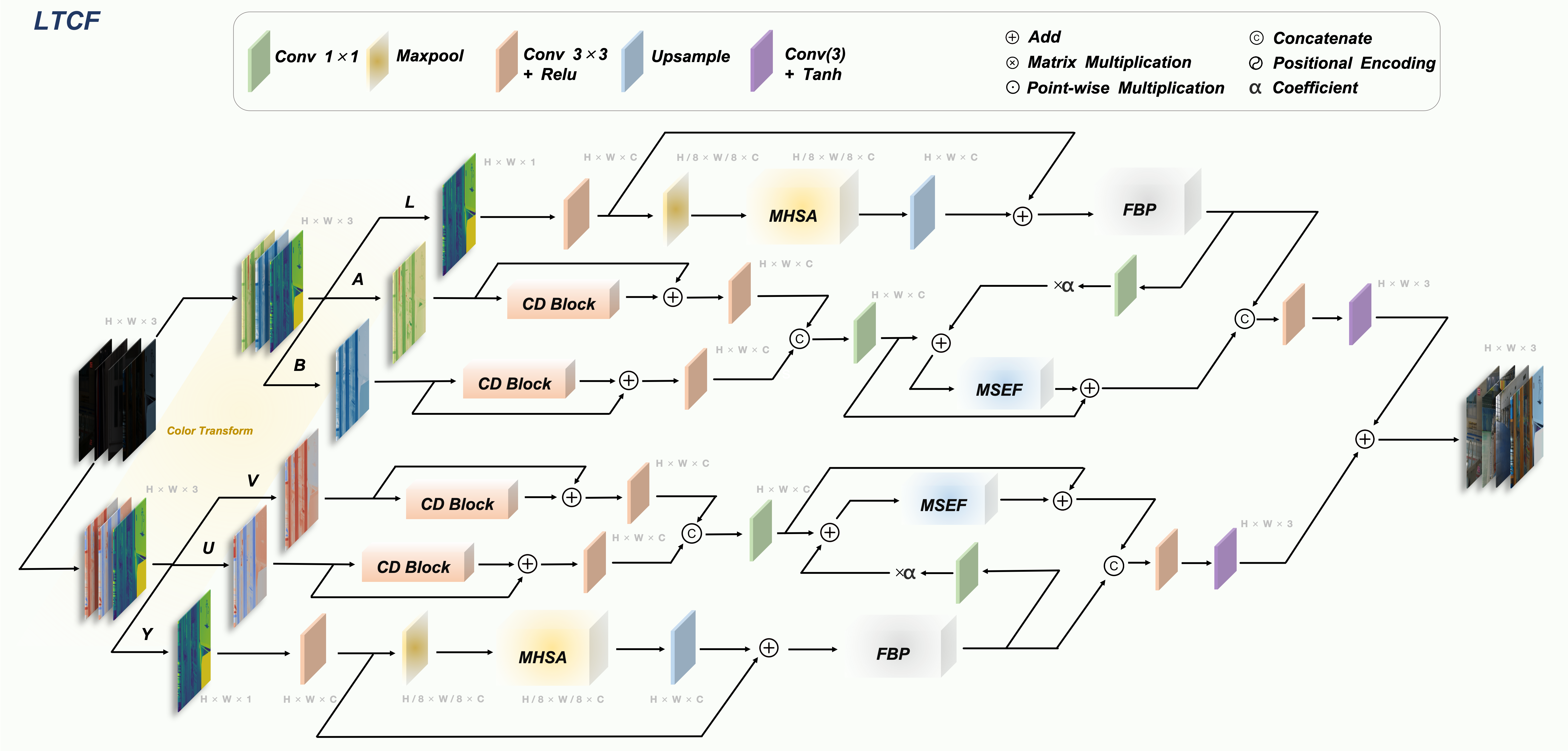}
    \caption{Model Pipeline. Our main models are Multi-header Self-attention (MHSA) Block, Channel Denoising (CD) Block, Multi-stage Squeeze and Excited Fusion (MSEF) Block and Fourier Branch Processing (FBP) Block. The individual submodules can be seen in Fig.2.}
    \label{fig1}
\vspace{-3mm}
\end{figure*}

To address these challenges, we present LTCF-Net, a novel approach to low-light image enhancement that leverages a dual-channel color space architecture. Our model uniquely processes illumination and color information in parallel streams, employing self-attention mechanisms~\cite{11} to dynamically adjust enhancement based on varying exposure levels across image regions. Unlike traditional Retinex-based methods~\cite{2,8} that rely on complex degradation modeling, LTCF-Net enables direct end-to-end training. By prioritizing illumination processing—aligned with human visual perception, our approach achieves superior detail preservation and natural enhancement while maintaining color fidelity, effectively avoiding common artifacts like overexposure and color distortion.

Experimental results demonstrate that our method outperforms existing state-of-the-art techniques on multiple datasets, achieving significant performance improvements on datasets such as LOL-v1 \cite{13}, LOL-v2 \cite{12}, SID \cite{15}, and SDSD \cite{14}, validating the effectiveness and superiority of our approach. Our main contributions are:

\begin{itemize} 
    \item We propose a novel dual-channel color space transformation method that effectively separates and independently processes illumination and color information, simplifying the complex decoupling task. The model utilizes LAB \cite{16} and YUV \cite{17} color spaces for targeted enhancement, implementing a multi-head self-attention mechanism \cite{11} on the luminance and chrominance layers to dynamically enhance low-light recovery capabilities.
    \item The Fourier adjustment module \cite{18,19} is introduced to convolve the real and imaginary parts of the data in the frequency domain to enhance the features in the frequency domain. It also includes an adaptive enhancement mechanism that balances the light distribution by dealing with regions of different scales or dynamic ranges
    \item Through quantitative and qualitative experiments, our network demonstrates superior performance on LOL \cite{12,13} and other datasets compared to state-of-the-art methods.
\end{itemize}

%% file: sec/3_Method.tex
\section{Proposed Method}

\begin{figure*}[ht] 
    \centering 
    \includegraphics[width=0.98\textwidth]{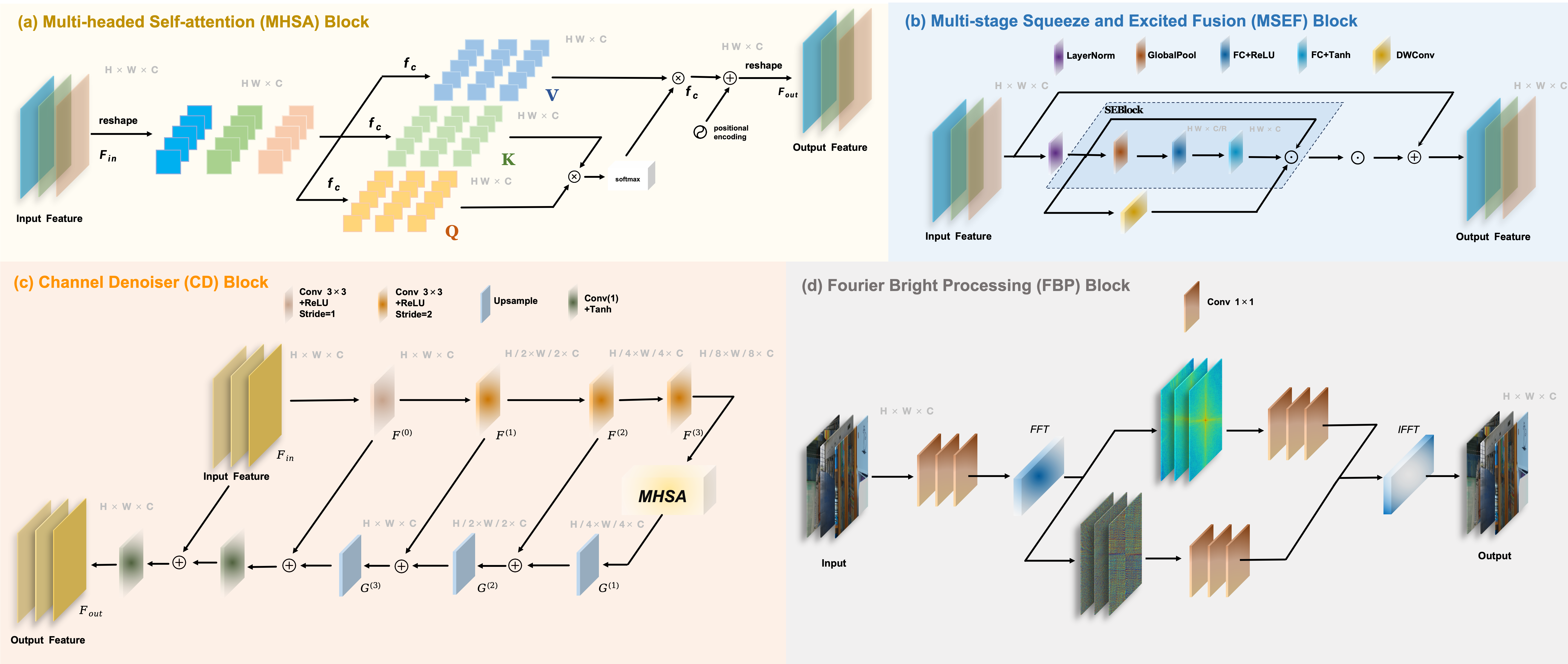} 
    \caption{Submodules in our model. (a) Multi-headed Self-attention (MHSA) Block use multi-head attention mechanism to acquire features. (b) Multi-stage Squeeze and Excited Fusion (MSEF) Block, Multi-stage processing and capture of global and local features for image recovery. (c) Channel Denoiser (CD) Block, step and jump connections based on U-shaped networks are used for denoising. (d) Fourier Bright Processing (FBP) Block uses Fourier transform to de-noise the light information.} 
    \label{fig2} 
\vspace{-3mm}
\end{figure*}

Fig.~\ref{fig1} and Fig.~\ref{fig2} illustrate the structure of the proposed LTCF-Net. As depicted in Fig.~\ref{fig1}, the architecture is divided into two branches dedicated to illumination enhancement, each operating within distinct color spaces: LAB and YUV. Within each branch, as illustrated in Fig.~\ref{fig2}(a) and Fig.~\ref{fig2}(d), a luminance enhancement channel incorporates both a Multi-Head Self-Attention (MHSA) module and a Fourier Brightness Processing (FBP) module, which are critical in meticulously restoring luminance details. These modules are detailed in Sections \ref{sec:3.2} and \ref{sec:3.5}, respectively.

Additionally, Fig.~\ref{fig2}(c) illustrates that each color channel integrates a Channel Denoising (CD) Block, outlined in Section \ref{sec:3.3}. This block enhances the noise reduction capabilities of the system. The denoised outputs are then combined with the luminance channel through a Multi-stage Squeeze and Excited Fusion (MSEF) module, elaborated in Section \ref{sec:3.4}. Fig.~\ref{fig2}(b) provides an in-depth view of the MSEF module, which is engineered to refine the modeling of both illumination and color features, thereby ensuring the fidelity and uniformity of the final enhanced images.

\subsection{Color Spaces}

The LAB and YUV color spaces offer significant advantages over the traditional RGB color space, particularly in terms of separating light information. This separation facilitates independent manipulation of the light and color channels, enhancing the process’s flexibility and effectiveness. The LAB color space, grounded in human visual perception, allows for precise adjustments as demonstrated by the standard LAB conversion formula presented in Eq.\ref{equ:1}.

\begin{align}
\begin{array}{r}
L^*=116 \times f\left(\frac{Y}{Y_n}\right)-16 \\
a^*=500\left[f\left(\frac{X}{X_n}\right)-f\left(\frac{Y}{Y_n}\right)\right] \\
b^*=200\left[f\left(\frac{Y}{Y_n}\right)-f\left(\frac{Z}{Z_n}\right)\right]
\end{array}
\label{equ:1}
\end{align}

Meanwhile, the YUV color space is particularly beneficial for low-light image enhancement due to its ability to independently adjust luminance (Y) — critical for enhancing visibility — without affecting the chrominance channels (U and V). The formulation of the YUV color space is specified in Eq.~\ref{equ:2}.

\begin{equation}
\begin{array}{r}
Y=0.299 R+0.587 G+0.114 B \\
U=-0.14713 R-0.28886 G+0.436 B \\
V=0.615 R-0.51499 G-0.10001 B
\end{array}
\label{equ:2}
\end{equation}

\subsection{Multi-header Self-attention Block}
\label{sec:3.2}
The MHSA Block, inspired by transformer architecture, begins by transforming the input feature $\mathbf{F}_{in} \in \mathbb{R}^{H \times W \times C}$ into a reshaped format $\mathbf{X} \in \mathbb{R}^{HW \times C}$. This reshaped feature is then partitioned into multiple ‘heads’ as $\mathbf{X}=\left[\mathbf{X}_1, \mathbf{X}_2, \cdots, \mathbf{X}_k\right]$.


Each segment, $\mathbf{X}_i \in \mathbb{R}^{HW \times d_k}$, where $d_k = \frac{c}{k}$ and $i = 1, 2, …, k$, undergoes processing by three bias-free fully connected layers ($f_c$), which project $\mathbf{X}_i$ into the query $\mathbf{Q}_i$, key $\mathbf{K}_i$, and value $\mathbf{V}_i$ components, which can be defined by $\mathbf{Q}_i=\mathbf{X}_i \mathbf{W}_{\mathbf{Q}_i}, \mathbf{K}_i=\mathbf{X}_i \mathbf{W}_{\mathbf{K}_i}, \mathbf{V}_i=\mathbf{X}_i \mathbf{W}_{\mathbf{V}_i}$.



The learnable parameters of the fully connected layers are represented by $\mathbf{W}_{\mathbf{Q}i}$, $\mathbf{W}_{\mathbf{K}i}$, and $\mathbf{W}_{\mathbf{V}_i} \in \mathbb{R}^{d_k \times d_k}$. This structure allows the model to adaptively respond to varying lighting conditions across different image regions, enhancing areas that are typically more challenging due to darkness. Each attention head functions independently, employing the self-attention mechanism as defined below:

\begin{equation}
    \text{Attention}(\mathbf{Q}_i,\mathbf{K}_i,\mathbf{V}_i)=\text{Softmax}\left(\frac{\mathbf{Q}_i\mathbf{K}_i^\mathrm{T}}{\sqrt{d_k}}\right)\times\mathbf{V}_i
\end{equation}

The outputs from all attention heads are concatenated and integrated using a positional encoder, then reshaped back to the original input dimensions, resulting in the final output feature $\mathbf{F}_{\mathrm{out}} \in \mathbb{R}^{H \times W \times C}$.

\subsection{Channel Denoising Block}
\label{sec:3.3}

The CD Block leverages a four-scale U-shaped network~\cite{44}, incorporating the MHSA mechanism introduced above at the network bottleneck. This integration of convolutional and attention-based methodologies allows for robust feature extraction and effective denoising. The module consists of multiple convolutional layers characterized by two types of strides and includes skip connections to enhance detail retrieval and noise reduction.

This block processes color information from two distinct color spaces, represented by $R_{i}$, where $i=4$ corresponds to the color space components $A$, $B$, $U$, and $V$. The process initiates with these color components being processed through a $3 \times 3$ convolutional layer with a stride of one to extract preliminary features, denoted by $F^{(0)} = \mathrm{Conv}_{3\times3}^{s=1}(R)$. The signal then sequentially traverses three $3 \times 3$ convolutional layers each with a stride of two, progressively capturing multi-scale features. For each convolution layer $k=1, 2, 3$, the operation results in $F^{(k)} = \mathrm{Conv}_{3\times3}^{s=2}(F^{(k-1)})$, reducing the spatial dimensions of the feature map by half with each convolution, transitioning from $F^{(1)} \in \mathbb{R}^{\frac{H}{2} \times \frac{W}{2} \times C}$ to $F^{(3)} \in \mathbb{R}^{\frac{H}{8} \times \frac{W}{8} \times C}$.

At the network bottleneck, minimum-scale feature mapping captures global dependencies effectively, utilizing the MHSA Block. Subsequent stages involve up-sampling, matched in scale to the prior down-sampling phases. The initial up-sampling step, employing a deconvolution with a stride of two, produces $G^{(1)} = \mathrm{Deconv}_{3\times3}^{s=2}(F^{(3)})$. This up-sampling process is repeated twice to restore the feature map to its original dimensions, $H \times W$. The final output is then refined through two additional convolutional layers followed by a Tanh activation function to maintain color fidelity. The ultimate output, $\mathbf{F}_{\mathrm{out}}$, restores the feature dimensions to $\mathbf{F}_{\mathrm{out}} \in \mathbb{R}_{i}^{H \times W \times C}$.

\subsection{Multi-stage Squeeze and Excited Fusion Block}
\label{sec:3.4}

The MSEF Block is engineered to integrate multiple feature enhancement mechanisms~\cite{45}, specifically designed to enhance the model’s capability to accurately reconstruct and refine details in images by capturing both global and local features associated with illumination and color information. Initially, the input feature is subjected to Layer Normalization to stabilize its mean and variance. Subsequently, the normalized features are processed through the Squeeze-and-Excitation Block (SEBlock), which employs a two-stage recalibration of feature importance: squeezing to reduce redundancy and excitation to emphasize informative features.

Within the SEBlock, features $\mathbf{F}_{in}$ initially pass through Global Average Pooling (GAP) to form a global descriptor for each channel, capturing essential contextual information that is broadly representative of the entire image. This descriptor $\mathbf{D}_{\mathrm{re}}$ is then refined through a fully connected layer equipped with a ReLU activation function, as outlined in Eq.~\ref{equ:7}, enhancing significant features while filtering out less relevant data. The process continues with another fully connected layer featuring a Tanh activation function, which expands the features back to their original dimensions to achieve optimal recalibration:

\begin{equation}
\begin{array}{r}
\mathbf{D}_{\mathrm{re}}=\operatorname{ReLU}\left(\mathbf{W}_1 \cdot \operatorname{GP}\left(\operatorname{LN}\left(\mathbf{F}_{\text {in }}\right)\right)\right) \\
\mathbf{D}_{\mathrm{ex}}=\operatorname{Tanh}\left(\mathbf{W}_2 \cdot \mathbf{D}_{\mathrm{re}}\right) \cdot \operatorname{LN}\left(\mathbf{F}_{\text {in }}\right)
\end{array}
\label{equ:7}
\end{equation}

This process not only emphasizes and restores key features but also enhances the model’s ability to detect variations in color and illumination, producing images that appear more natural. This is particularly vital for low-light image enhancement, where maintaining accurate color and detail is crucial. Finally, residual connections also help preserve the original input features, ensuring stability during training and preventing the loss of important details.


\begin{figure*}[ht] 
    \centering 
    \includegraphics[width=0.98\textwidth]{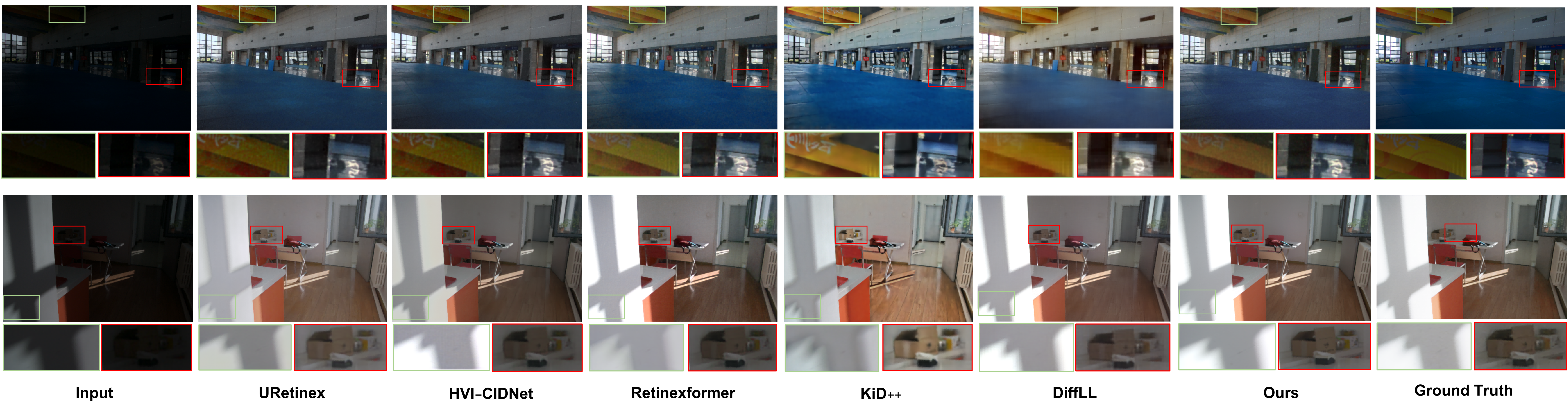} 
    \vspace{-2mm}
    \caption{Results on LOL-v1\cite{13}(top) and LOL-v2-real\cite{12}(bottom). Our method effectively enhances the visibility and preserves the color.} 
    \label{fig3} 
    \vspace{-3mm}
\end{figure*}

\subsection{Fourier Branch Processing Block}
\label{sec:3.5}
The Fourier Branch Processing Block is crucial for enhancing and restoring image brightness by manipulating the luminance channel in the frequency domain through the application of Fourier Transform techniques~\cite{46}. This approach allows for precise differentiation and manipulation of high-frequency and low-frequency components, thereby significantly improving the details in brightness across the image.


Initially, the luminance channel undergoes transformation into its frequency-domain representation, with the real and imaginary parts processed separately to refine frequency-domain features. A convolutional layer reduces the number of channels to 16, streamlining features while preserving essential details. This is followed by a Leaky ReLU activation function that captures finer details and mitigates the issue of dying ReLUs in areas of low brightness. Subsequently, features are further refined through an additional convolutional layer that maintains the channel count, ensuring that critical nuances are preserved. A final convolutional layer re-expands the features to match the original input dimensions. This restoration is crucial for aligning the enhanced features with the original image structure, facilitating seamless integration back into the overall image processing workflow. Detailed experimental validations of this process are presented in Section~\ref{sec:3.4}.


\subsection{Combined Loss Function}
In this section, we introduce our loss function framework, dividing six loss variables into two main categories: pixel-level and perceptual-level losses, each targeting different aspects of image quality enhancement.

\noindent\textbf{Pixel-Level Losses.} These losses measure discrepancies directly at the pixel level, thus preserving low-level image attributes such as brightness and color fidelity. 

Specifically, $\mathcal{L}_{\mathrm{S1}}$ is a smooth L1 loss function, which is defined for two parameters: the predicted image ($y_{\mathrm{pred}}$) and the ground truth image ($y_{\mathrm{true}}$). This loss function has been proven effective in Fast R-CNN because it is robust to outliers and can provide stable gradients during the training process.  Its formula is shown in Eq.~\ref{equ:loss1}
\begin{align}
\mathcal{L}_{\text{S1}}(y_{\mathrm{true}}, y_{\mathrm{pred}}) &= \sum_{i \in \{x, y, w, h\}} \text{smooth}_{L_1}(y_{\mathrm{true}}^i - y_{\mathrm{pred}}^i) \label{equ:loss1}
\end{align}
Where the function $smooth_{L_1(x)}$is defined as:
\begin{equation}
\begin{array}{r}
smooth_{L_1(x)} = \begin{cases}
0.5 x^2, & \text{if } |x| < 1 \\
|x| - 0.5, & \text{otherwise}
\end{cases}
\end{array}
\end{equation}

\begin{table*}[t]
	\centering
	\setlength\tabcolsep{4pt}
	\resizebox{0.995\textwidth}{!}{\hspace{-0.5mm}
		\begin{tabular}{
        >{\centering\arraybackslash}p{3cm}| 
        >{\centering\arraybackslash}p{1.2cm}>{\centering\arraybackslash}p{1.2cm}| 
        >{\centering\arraybackslash}p{1.2cm}>{\centering\arraybackslash}p{1.2cm}| 
        >{\centering\arraybackslash}p{1.2cm}>{\centering\arraybackslash}p{1.2cm}| 
        >{\centering\arraybackslash}p{1.2cm}>{\centering\arraybackslash}p{1.2cm}| 
        >{\centering\arraybackslash}p{1.2cm}>{\centering\arraybackslash}p{1.2cm}| 
        >{\centering\arraybackslash}p{1.2cm}>{\centering\arraybackslash}p{1.2cm}| 
        >{\centering\arraybackslash}p{1.7cm}>{\centering\arraybackslash}p{1.7cm} 
    }
			\toprule[0.15em]
			\multirow{2}{*}{Methods} & \multicolumn{2}{c|}{LOL-v1} & \multicolumn{2}{c|}{LOL-v2-real} & \multicolumn{2}{c|}{SID} & \multicolumn{2}{c|}{SMID} & \multicolumn{2}{c|}{SDSD-indoor} & \multicolumn{2}{c|}{SDSD-outdoor} & \multicolumn{2}{c}{Complexity} \\ 
			& PSNR & SSIM & PSNR & SSIM & PSNR & SSIM & PSNR & SSIM & PSNR & SSIM & PSNR & SSIM & FLOPS (G) & Params (M) \\ \midrule[0.15em]
			SID\cite{15} & 14.35 & 0.43 & 13.24 & 0.44 & 16.97 & 0.59 & 24.78 & 0.71 & 23.29 & 0.70 & 24.90 & 0.69 & 13.43 & 7.76 \\			
			DeepUPE\cite{59} & 14.38 & 0.44 & 13.27 & 0.45 & 17.01 & 0.60 & 23.91 & 0.69 & 21.70 & 0.66 & 21.94 & 0.69 & 21.10 & 1.02 \\
			IPT\cite{65} & 16.27 & 0.50 & 19.80 & 0.83 & 20.53 & 0.56 & 27.03 & 0.78 & 26.11 & 0.83 & 27.55 & 0.85 & 6887 & 115.31 \\
			RetinexNet\cite{13} & 16.77 & 0.56 & 15.47 & 0.56 & 16.48 & 0.57 & 22.83 & 0.68 & 20.84 & 0.61 & 20.96 & 0.62 & 587.47 & 0.84 \\ 
			EnGAN\cite{4} & 17.48 & 0.65 & 18.23 & 0.61 & 17.23 & 0.54 & 22.62 & 0.67 & 20.02 & 0.60 & 20.10 & 0.61 & 61.01 & 114.35 \\ 
			FIDE~\cite{61} & 18.27 & 0.66 & 16.85 & 0.67 & 18.34 & 0.57 & 24.42 & 0.69 & 22.41 & 0.65 & 22.20 & 0.62 & 28.51 & 8.62 \\
			DRBN~\cite{62} & 20.13 & 0.83 & 20.29 & 0.83 & 19.02 & 0.57 & 26.60 & 0.78 & 24.08 & 0.86 & 25.77 & 0.84 & 48.61 & 5.27 \\
			KinD++~\cite{35} & 20.86 & 0.79 & 14.74 & 0.64 & 18.02 & 0.58 & 22.18 & 0.63 & 21.95 & 0.67 & 21.97 & 0.65 & 34.99 & 8.02 \\
			URetinex~\cite{37} & 21.32 & 0.83 & 22.79 & 0.83 & 19.07 & 0.61 & 22.47 & 0.67 & 24.88 & 0.73 & 24.96 & 0.76 & 58.27 & 0.36 \\ 
			MIRNet~\cite{36} & 24.14 & 0.83 & 20.02 & 0.82 & 20.84 & 0.60 & 25.66 & 0.76 & 24.38 & \cellcolor{yellow!30}0.86 & 27.13 & 0.83 & 785 & 31.76 \\   
			SNR-Net~\cite{40} & 24.61 & 0.84 & 21.48 & 0.84 & 22.87 & 0.62 & 28.49 & 0.80 & 29.44 & \cellcolor{red!20}0.89 & 28.66 & \cellcolor{yellow!30}0.86 & 26.35 & 4.01 \\
			Retinexformer~\cite{8} & 25.16 & 0.84 & 22.80 & 0.84 & 24.44 & \cellcolor{yellow!30}0.68 & \cellcolor{blue!10}29.15 & \cellcolor{yellow!30}0.81 & \cellcolor{yellow!30}29.77 & \cellcolor{red!20}0.89 & \cellcolor{blue!10}29.84 & \cellcolor{red!20}0.87 & 15.57 & 1.61 \\ 
			DiffLL~\cite{47} & 26.33 & 0.83 & \cellcolor{blue!10}28.85 & \cellcolor{blue!10}0.87 & \cellcolor{yellow!30}25.88 & 0.66 & 28.71 & 0.72 & 27.92 & 0.79 & 28.82 & 0.80 & 124.48 & 0.66 \\
			LYT-Net~\cite{60} & \cellcolor{yellow!30}27.23 & \cellcolor{blue!10}0.85 & 27.80 & \cellcolor{blue!10}0.87 & 24.19 & 0.62 & 28.73 & 0.73 & 29.20 & 0.77 & \cellcolor{yellow!30}29.87 & 0.82 & 3.49 & 0.045 \\
			HVI-CIDNet~\cite{49} & \cellcolor{red!20}27.71 & \cellcolor{yellow!20}0.87 & 28.13 & 0.89 & 22.90 & \cellcolor{blue!10}0.67 & 28.63 & \cellcolor{blue!10}0.79 & \cellcolor{blue!10}29.54 & \cellcolor{yellow!30}0.86 & 27.54 & 0.79 & 7.57 & 1.88 \\  
			\midrule[0.15em]
			\textbf{LTCF-Net*} & 26.88 & \cellcolor{red!20}0.89 & \cellcolor{red!20}30.11 & \cellcolor{red!20}0.93 & \cellcolor{blue!10}25.26 & \cellcolor{yellow!30}0.68 &  \cellcolor{yellow!30}29.58 & \cellcolor{red!20}0.86 & 29.27 & \cellcolor{blue!10}0.80 & 28.70 & \cellcolor{blue!10}0.84 & 4.60 & 0.097 \\ 
			\textbf{LTCF-Net} & \cellcolor{blue!10}27.07 & \cellcolor{red!20}0.89 & \cellcolor{yellow!30}29.76 & \cellcolor{yellow!30}0.92 & \cellcolor{red!20}26.28 & \cellcolor{red!20}0.70 & \cellcolor{red!20}29.63 & \cellcolor{red!20}0.86 & \cellcolor{red!20}29.98 & \cellcolor{red!20}0.89 & \cellcolor{red!20}30.14 & \cellcolor{red!20}0.87 & 10.370 & 0.155 \\ \bottomrule[0.15em]
		\end{tabular}}

	\vspace{2mm}
	\caption{Quantitative comparisons on LOL (v1~\cite{13} and v2~\cite{12}), SID~\cite{15}, SMID~\cite{48}, and SDSD~\cite{14} (indoor and outdoor) datasets. The best result is in \textcolor{red!20}{red}, the second best result is in \textcolor{yellow}{yellow} and the third best result is in in \textcolor{blue!20}{blue}. Our LTCF-Net significantly outperforms SOTA algorithms. LTCF-Net* represents the proposed method without the Fourier Branch Processing Block.}
	\label{tab:quantitative}
	\vspace{-3mm}
\end{table*}

$\mathcal{L}_{\mathrm{PSNR}}$ measures and optimizes image quality by assessing the mean squared error between (\(y_{\text{pred}}\)) and (\(y_{\text{true}}\)). It is designed to enhance pixel accuracy, as detailed in Eq.~\ref{equ:losspsnr}.


\begin{align}
\mathcal{L}_{\text{PSNR}} = 40.0 - \sum_{i=1}^{n} 20 \cdot \log_{10}\left(\frac{1}{\sqrt{\text{MSE}}}\right)
\label{equ:losspsnr}
\end{align}

The $\mathcal{L}_{\mathrm{Color}}$ ensures color fidelity by minimizing color discrepancies between the predicted and reference images, calculated as shown in Eq.~\ref{equ:losscolor}. 

\begin{align}
\mathcal{L}_{\text{Color}} =  \sum_{i=1}^{n}\left| \text{mean}(y_{\text{true}}^i) - \text{mean}(y_{\text{pred}}^i) \right|
\label{equ:losscolor}
\end{align}

The histogram loss function aligns the global statistical features such as brightness and contrast by comparing histograms, enhancing overall image consistency as illustrated in Eq.~\ref{equ:lossHist}.

\begin{align}
\mathcal{L}_{\text{Hist}} =  \sum_{i=1}^{n} \left| \frac{1}{N} \sum_{n=1}^{N} H_{\text{true}}^{i,n} - \frac{1}{N} \sum_{n=1}^{N} H_{\text{pred}}^{i,n} \right|
\label{equ:lossHist}
\end{align}

Where \(H_{\text{true}}^{i,n}\) and \(H_{\text{pred}}^{i,n}\) are the histogram values for the ground truth and predicted images, respectively, at bin \(n\). This loss encourages the model to generate predictions whose color distributions match those of the ground truth more closely.

\noindent\textbf{Perceptual-Level Losses.} These losses, on the other hand, leverage high-level features to enhance the visual quality of generated images. 

$\mathcal{L}_{\mathrm{per}}$ utilizes features extracted from a VGG19 network, this perceptual loss measures differences that affect the perceptual quality between the predicted and ground truth images. Its formula is shown in Eq.~\ref{equ:lossperc}.
\begin{equation}
\begin{array}{r}
\mathcal{L}_{\text{per}} = \frac{1}{C_j H_j W_j} \left\| \phi_j(y_{true}) - \phi_j(y_{pred}) \right\|_2^2
\end{array}
\label{equ:lossperc}
\end{equation}

\begin{figure*}[ht] 
    \centering 
    \includegraphics[width=0.98\textwidth]{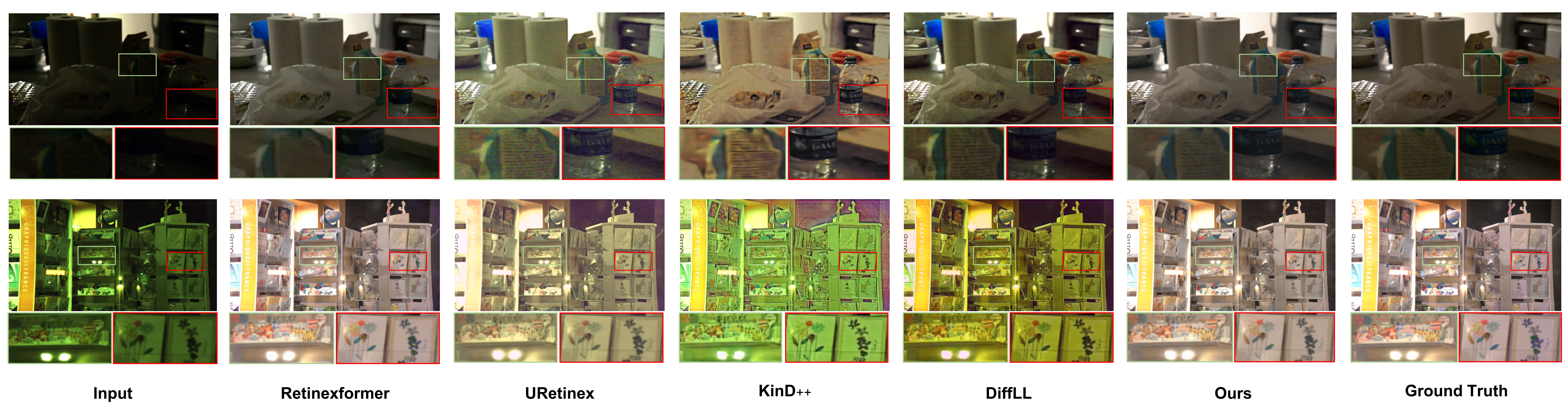} 
    \vspace{-2mm}
    \caption{Results on SID\cite{15}(top) and SMID\cite{48}(bottom). Previous methods either output incorrect colors or have strong noise. Our method effectively improves the visibility of the image and preserves the rich details.} 
    \label{fig4} 
\vspace{-4mm}
\end{figure*}

\begin{figure*} 
    \centering 
    \includegraphics[width=0.98\textwidth]{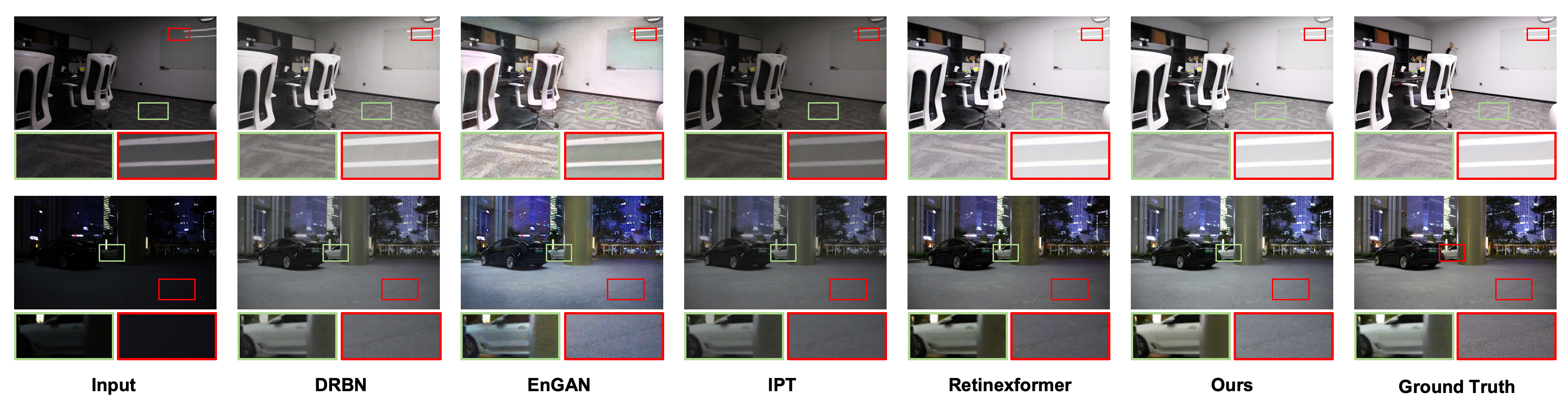} 
\vspace{-3mm}    
\caption{Results on SDSD-indoor~\cite{14}(top) and SDSD-outdoor~\cite{14}(bottom). Benchmark methods either output incorrect colors or have strong noise, while the proposed method effectively improves the visibility of the image and preserves the rich details.} 
    \label{fig6}
\vspace{-3mm}
\end{figure*}

Additionally, $\mathcal{L}_{\mathrm{SSIM}}$ is based on multi-scale structural similarity, this loss ensures structural consistency across different scales, thereby enhancing the perceptual similarity of the generated image. Its formula is shown in Eq.~\ref{equ:lossssim}. Where \( y \) is the original image and \( \hat{y} \) is the target image.

\begin{equation}  
\mathcal{L}_{\text{SSIM}}(y,\hat{y}) = 1 - \frac{(2 \mu_y \mu_{\hat{y}} + C_1)(2 \sigma_{y\hat{y}} + C_2)}{(\mu_y^2 + \mu_{\hat{y}}^2 + C_1)(\sigma_y^2 + \sigma_{\hat{y}}^2 + C_2)}
\label{equ:lossssim}
\end{equation}

As shown in Eq~\ref{equ:total}, the final loss function is a combined loss between pixel level losses and perceptual level losses, where $\alpha_1$ to $\alpha_5$ are hyperparameters for each loss component. This final loss function includes provides a comprehensive evaluation metric that optimizes various aspects of image quality, ultimately leading to a more realistic and high-quality output.

\begin{equation}
\begin{aligned}
\mathcal{L}_{Pixel} &= \mathcal{L}_{\mathrm{S1}} + \alpha_1 \mathcal{L}_{\mathrm{PSNR}}+ \alpha_2 \mathcal{L}_{\mathrm{Color}}+ \alpha_3 \mathcal{L}_{\mathrm{Hist}}   \\
\mathcal{L}_{Perc} &= \alpha_4 \mathcal{L}_{\mathrm{Per}} + \alpha_5 \mathcal{L}_{\mathrm{SSIM}} \\
\mathcal{L}_{Total} &= \mathcal{L}_{Pixel} + \mathcal{L}_{Perc}
\label{equ:total}
\end{aligned}
\end{equation}

%% file: sec/4_Experiment.tex
\section{Experiment}
\subsection{Datasets and Implementation Details}
We eveluate our method on LOL(v1~\cite{13} and v2-real~\cite{12}), SID~\cite{15}, SMID~\cite{48}, SDSD~\cite{14}, and FiveK~\cite{50} datasets.

\noindent\textbf{LOL.} We use both LOL-v1 and LOL-v2-real dataset. The training and testing sets are split with ratios of 485:15, 689:100 for LOL-v1, LOL-v2-real respectively.

\noindent\textbf{SID.} A subset of the SID dataset captured using the Sony $\alpha7SII$ camera is used for evaluation. This subset consists of 2697 pairs of short- and long-exposure RAW images. Low light and normal light RGB images are generated by applying the same in-camera signal processing as used in SID to convert RAW images to RGB. Of these pairs, 2299 are allocated for training and 398 for testing.

\noindent\textbf{SMID.} The SMID benchmark dataset contains 20809 pairs of short- and long-exposure RAW images. These RAW images are also converted to low-light and normal-light RGB image pairs. The dataset provides 18789 pairs for training and 2021 pairs for testing.

\noindent\textbf{SDSD.} We use the static version of the SDSD dataset captured with a Canon EOS 6D Mark $II$ camera and an ND filter. The SDSD dataset includes both indoor and outdoor subsets, with the SDSD-outdoor subset containing 3150 images and the SDSD-indoor subset containing 1963 images.

\noindent\textbf{FiveK.} FiveK dataset is adjusted by experts as references to create low-light photos followed by the method proposed in~\cite{50}. It contains 5000 underexposed image pairs with bounding box annotations for 60 object categories. Note that this dataset is particularly used in Object Detection task for enhanced low-light images as discussed in Section~\ref{sec:3.3}.

In addition to the above eight benchmarks, we also tested our method on five datasets: LIME~\cite{51}, NPE~\cite{52}, MEF~\cite{53}, DICM~\cite{54}, and VV~\cite{55}, which do not have ground truth annotations.

\begin{table*}[t]\hspace{0mm}
    \subfloat[\scriptsize User surveys score in each algorithm.\label{user_study}]{\vspace{2mm}
        \setlength{\tabcolsep}{1.3mm}
        \scalebox{0.63}{
            \centering
            \begin{tabular}{lccccccc}
                \toprule[0.15em]
                Methods &L-v1 &L-v2 &SID  &SMID  &SD-in  &SD-out  &Mean\\
                \midrule[0.15em]
                
                HVI-CIDNet & 2.97 & 3.27 & 2.75 & 3.23 & 3.34 & 3.48 & 3.17  \\
                
                Retinexformer & \textbf{3.52$^{*}$} & \textbf{4.01$^{*}$} & \textbf{3.19$^{*}$} & 3.75 & 3.66 & \textbf{3.98$^{*}$} & \textbf{3.69$^{*}$}\\
                
                URetinex &2.81 &3.01 &2.69 &\textbf{3.88$^{*}$} &3.75 &3.82 &3.33 \\
                
                LYT-Net &3.6 &3.77 &3.1 &3.72 &3.77 &3.83 &3.63\\
                
                GASD &2.99 &3.21 &2.88 &3.02 &\textbf{\(\uparrow\)3.84} &3.24 &3.20 \\
                
                KID++ &2.5 &2.64 &2.78 &3.21 &3.01 &3.22 &2.90 \\
                
                DIffLL &2.65 &2.72 &2.69 &3.01 &3.27 &3.08 &2.91 \\
                
                \midrule[0.15em]
                \textbf{LTCF-Net}   &\textbf{\(\uparrow\)3.69} &\textbf{\(\uparrow\)4.23} &\textbf{\(\uparrow\)3.21} &\textbf{\(\uparrow\)4.11} & \textbf{3.78$^{*}$} &\textbf{\(\uparrow\)4.01} &\textbf{\(\uparrow\)3.84}  \\ 
                \bottomrule[0.15em]
            \end{tabular}}}\vspace{0mm}
   \hspace{2mm}
    \subfloat[\scriptsize Average accuracy of object detection by different algorithms in FiveK dataset. \label{exdark}]{\vspace{2mm}
        \setlength{\tabcolsep}{1.3mm}
        \scalebox{0.63}{
            \centering
            \begin{tabular}{lccccccccccccc}
                \toprule[0.15em]
                Methods &Bicycle &Boat &Bottle &Bus &Car &Cat &Chair &Cup &Dog &Motor &People &Table &Mean\\
                \midrule[0.15em]
                
                HVI-CIDNet &41.3 &31.4 &32.5 &46.3 &44.5 &26.4 &25.5 &26.5 &27.4 &25.5 &31.4 &25.5 &32.1\\
                
                Retinexformer &44.5 &33.8 &32.5 &48.5 &46.8 &28.7 &27.8 &28.3 &29.1 &28.9 &32.7 &26.8 &34.2  \\
                
                URetinex &\textbf{46.2$^{*}$} & \textbf{36.2$^{*}$} &35.6 &51.2 &49.2 &\textbf{\(\uparrow\)30.2} &\textbf{30.3$^{*}$} &30.7 &32.3 &31.2 &34.2 &28.2 &\textbf{36.2$^{*}$}  \\
                
                LYT-Net &34.5 &34.5 &31.8 &47.6 &45.7 &27.6 &26.7 &27.8 &26.8 &27.3 &30.8 &25.8 &32.4  \\
                
                GASD &43.8 &32.7 &34.1 &49.8 &47.3 &\textbf{29.1$^{*}$} &28.1 &\textbf{31.4$^{*}$} &30.9 &\textbf{33.1$^{*}$} &35.3 &27.7 &35.3  \\
                
                KID++ &40.6 &35.1 &\textbf{\(\uparrow\)36.4} &\textbf{52.4$^{*}$} &\textbf{\(\uparrow\)50.1} &25.8 &29.6 &\textbf{\(\uparrow\)33.2} &\textbf{33.6$^{*}$} &29.7 &37.1 &\textbf{29.1$^{*}$} &36.1 \\
                
                DIffLL &45.1 &30.9 &30.7 &45.7 &43.9 &27.3 &\textbf{\(\uparrow\)30.9} &25.6 &25.7 &32.8 &\textbf{38.6$^{*}$} &26.9 &33.7 \\
                \midrule[0.15em]
                \textbf{LTCF-Net} &\textbf{\(\uparrow\)47.8} &\textbf{\(\uparrow\)36.9} &\textbf{36.2$^{*}$} &\textbf{\(\uparrow\)53.4} &\textbf{49.8$^{*}$} &28.5 &29.7 &31.2 &\textbf{\(\uparrow\)34.6} &\textbf{\(\uparrow\)33.9} &\textbf{\(\uparrow\)39.7} &\textbf{\(\uparrow\)30.1} &\textbf{\(\uparrow\)37.7} \\ 
                \bottomrule[0.15em]
            \end{tabular}}}\vspace{0mm}
   \vspace{-3mm}
    \caption{\small (a) compares the human perception quality of various low-light enhancement algorithms conduced by user survey. (b) compares the preprocessing effects of different methods on high-level vision understanding. Our model LTCF-Net outperforms the benchmarks on average.}
    \label{tab:us_lod}\vspace{-4mm}
\end{table*}

\begin{figure*}[ht] 
\centering 
\includegraphics[width=0.98\textwidth]{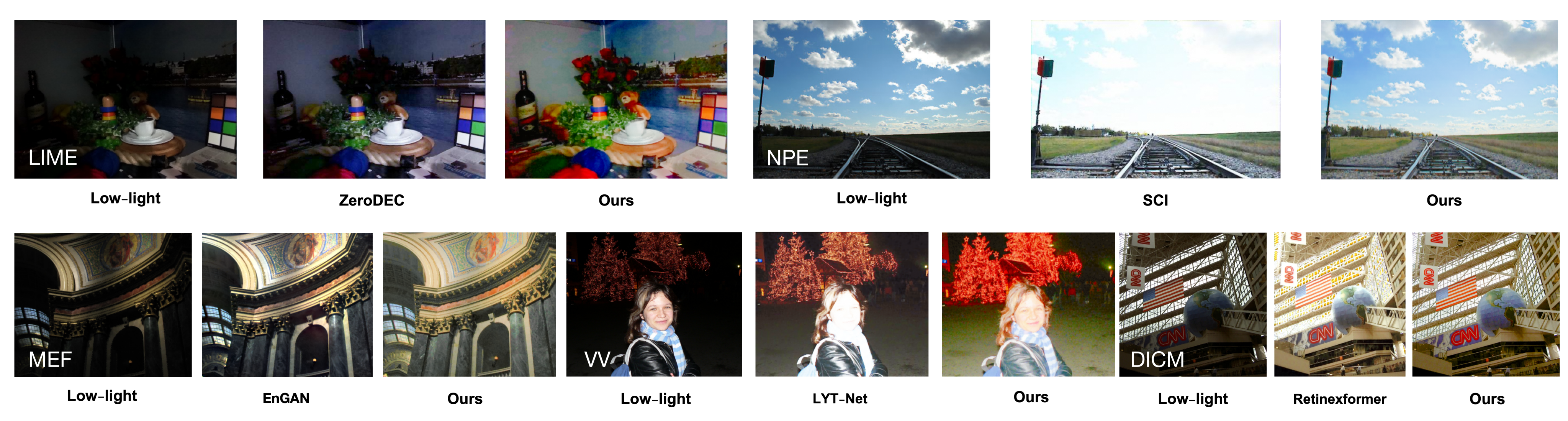} 
\vspace{-2mm}
\caption{Visual results on the LIME~\cite{51}, NPE~\cite{52}, MEF~\cite{53}, DICM~\cite{54}, and VV~\cite{55} datasets. Our LTCF performs better.} 
\label{fig5} 
\vspace{-2mm}
\end{figure*}

\noindent\textbf{Implementation Details.} Our model is implemented using the PyTorch framework and trained with the ADAM optimizer~\cite{56}, using parameters $\beta_{1} = 0.9$ and $\beta_{2} = 0.999$, for 1000 epochs. Training and testing process were conducted on a server equipped with an NVIDIA A40 GPU. The network parameters were randomly initialized, and the model was trained from scratch. An initial learning rate of $2 \times 10^{-4}$ was set and gradually reduced to $1 \times 10^{-6}$ using a cosine annealing scheduler~\cite{57} to facilitate convergence and avoid local minima. For the hyperparameters in the loss function, we set $\alpha_{1} = 0.12$, $\alpha_{2} = 0.05$, $\alpha_{3} = 0.55$, $\alpha_{4} = 0.015$, and $\alpha_{5} = 0.25$. 

\vspace{-3mm}
\subsection{Comparison Study of Low-light Image Enhancement}

\noindent\textbf{Quantitative Results.} The proposed method is quantitatively evaluated against various SOTA algorithms, as detailed in Table~\ref{tab:quantitative}. While our model registers a slightly lower PSNR on the LOL-v1 dataset, it shows a notable improvement of 1.26 dB in PSNR on the LOL-v2-real dataset compared to the leading DiffLL model. Additionally, it surpasses other top-performing methods on the SID, SMID, SDSD-indoor, and SDSD-outdoor datasets with respective PSNR enhancements of 0.4 dB, 0.43 dB, 0.21 dB, and 0.27 dB. This performance is achieved with only 10.37G FLOPS and 0.155M Params, demonstrating the efficiency and effectiveness of our approach.

\noindent\textbf{Qualitative Results.} As shown in Fig.~\ref{fig3}, we first compare the visual results of our model against others across the LOL-v1 and LOL-v2-real datasets. For instance, Retinexformer~\cite{8} and URetinex~\cite{37} demonstrate subpar performance in restoring light and shadow details. Similarly, DiffLL~\cite{47} and HVI-CIDNet~\cite{49} struggle with accurate color restoration. In contrast, our model excels in accurately enhancing both light and intricate details. 

Additionally, as depicted in Fig.~\ref{fig4}, models like URetinex \cite{37}, KinD++ \cite{35}, and DiffLL \cite{47} show noticeable color inaccuracies. Moreover, Retinexformer \cite{8} is unable to recover text on a sticker, underscoring its limitations. These observations confirm that our method outperforms others in restoring contrast and detail, particularly in the SID and SMID datasets. Including Fig.~\ref{fig6}, in SDSD-indoor and outdoor datasets, our method can restore many details.

\noindent\textbf{User Study Score.} We conducted a user study involving 95 participants to subjectively evaluate the visual quality of images from seven different datasets as shown in Fig.~\ref{fig5}. Each participant evaluated the enhanced images based on three criteria: overall reconstruction quality, presence of overexposure, and accuracy of detailed recovery. Each participant rated the enhanced images on a scale from 0 to 5. According to the aggregated scores presented in Table~\ref{tab:us_lod}(a), our method consistently outperformed competing methods, effectively avoiding issues of overexposure and underexposure while ensuring uniformly distributed brightness across the images.


\subsection{Low-light Object Detection Comparisons}
\label{sec:3.3}
Additional, we evaluate the impact of various enhancement algorithms on object detection using the FiveK dataset~\cite{50}. We used YOLO-v4-tiny~\cite{58} as the detector, and then compare the object detection performance on raw images and the images enhanced by different low-light enhancement methods.

Table~\ref{tab:us_lod} shows the average precision (AP) scores. Our model achieves the highest mean AP of 37.7, surpassing the leading self-monitoring method, URetinex, by 1.5 AP. Notably, our approach outperforms in seven object categories: Bicycle, Boat, Bus, Dog, Motor, People, and Table.

\begin{figure}
\vspace{-2mm}
	\includegraphics[width=0.9\linewidth]{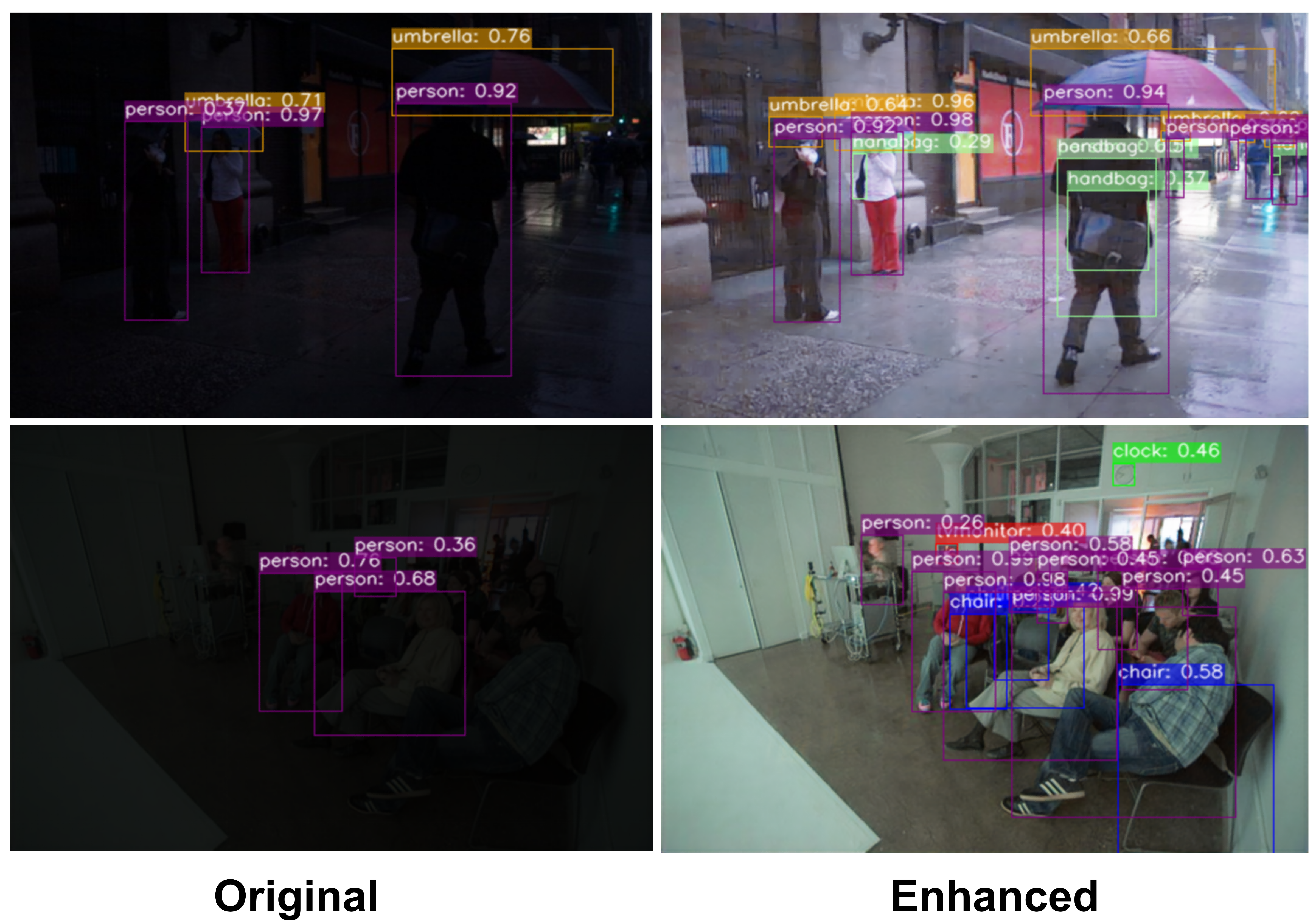}
	\caption{Visual comparison of object detection in low-light (left) and enhanced (right) scenes on the FiveK\cite{50} dataset.}
 \label{fig7} 
\vspace{-5mm}
\end{figure}

Fig.~\ref{fig7} illustrates the qualitative detection results in a low-light scene (left) and after LTCF-Net enhancement (right). The enhanced image enables the detector to identify more objects with higher accuracy, demonstrating the crucial role of our low-light enhancement method in improving object detection performance.

\begin{figure}
\vspace{0mm}
	\includegraphics[width=1\linewidth]{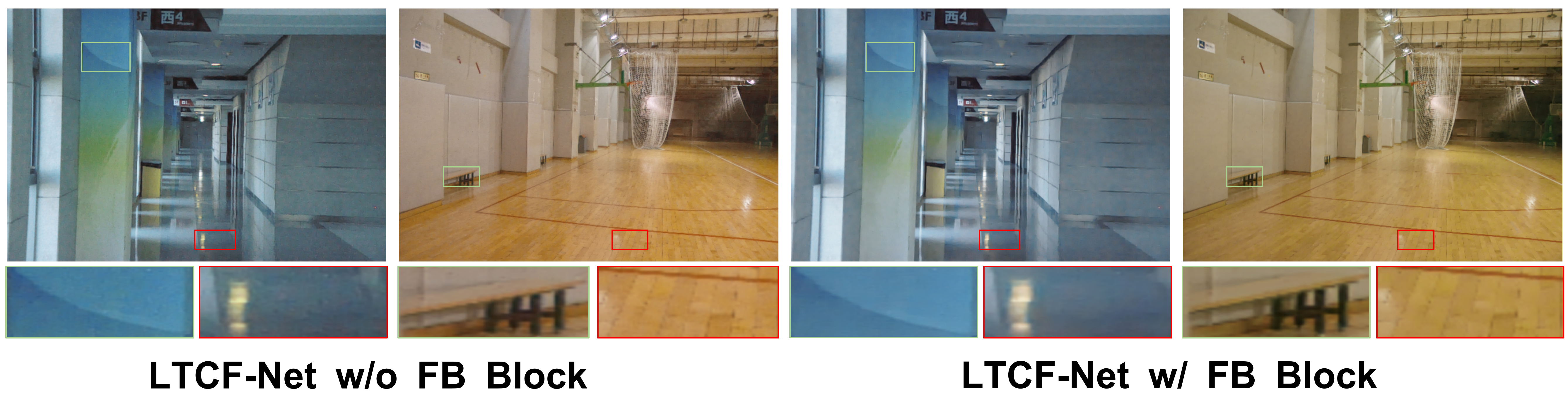}
	\caption{A detailed comparison between our two models shows the model without the Fourier module on the left and the model with the Fourier module on the right.}
 \label{fig8} 
 \vspace{-4mm}
\end{figure}


\subsection{Ablation Study}
\label{sec:3.4}
To validate the effectiveness of each component within the LTCF-Net and its contribution to training optimization, we conducted detailed ablation experiments on the SID dataset, specifically examines the dual color branches, the MSEF module, and the FBP module.

\noindent\textbf{Effectiveness of Dual Color Space Branch.} We assessed the impact of individual color spaces on model performance through a series of decomposed experiments, as detailed in Table~\ref{table3}. By separately eliminating the LAB and YUV color spaces, and comparing these to the dual color space setup, we observed the effect on performance metrics. The results indicate that the combined use of both color spaces significantly enhances performance over any single color space configuration. 

\noindent\textbf{Effectiveness of MSEF Block.} The incorporation of the MSEF module, despite increasing the number of parameters and FLOPS, results in significant improvements in image quality metrics, boosting the metric of PSNR by 1.38 and SSIM by 0.08. These gains confirm that the inclusion of the MSEF module is crucial for enhancing the model’s performance.

\begin{table}[]
\centering
\scalebox{0.7}{
    \begin{tabular}{c c c  c c c c c}
        \toprule
        LAB & YUV  & MSEF & FBP  & PSNR & SSIM & Params (M) & FLOPS (G) \\
        \midrule
        \checkmark  &           &           &            & 22.07 & 0.588 & 0.044 & 3.38 \\
                    & \checkmark &           &           & 22.12 & 0.587 & 0.042 & 3.33 \\
        \checkmark  &           & \checkmark &           & 24.08 & 0.610 & 0.047 & 3.55 \\
                    & \checkmark & \checkmark &          & 24.19 & 0.622 & 0.045 & 3.50 \\
        \checkmark  & \checkmark &           &           & 23.88 & 0.601 & 0.078 & 4.23 \\
        \checkmark  & \checkmark & \checkmark &          & 25.26 & 0.682 & 0.097 & 4.60 \\
        \checkmark  & \checkmark & \checkmark & \checkmark &\bf 26.28 &\bf 0.701 & 0.155 & 10.37 \\
        \bottomrule
    \end{tabular}}\vspace{0mm}
    \caption{\small Ablation study on each proposed network component tested on the SID~\cite{15} dataset, with PSNR, SSIM, params, FLOPS(size=256$\times$256) metrics reported.}
     \label{table3} 
     \vspace{-5mm}
\end{table}

\noindent\textbf{Effectiveness of FBP Block.} The primary function of this module is to reduce noise across the image. Observations from Table~\ref{table3} and Fig.~\ref{fig8} indicate that models lacking the Fourier module exhibit noticeable noise. Integrating Fourier modules significantly reduce this noise, albeit at the cost of slightly increasing the overall brightness of the image. Given that the final assessment of image quality relies on visual perception, the inclusion of Fourier modules is deemed essential. Consequently, we offer two versions of the model to accommodate different preferences and requirements.

%% file: sec/5_Conclusion.tex
\vspace{-2mm}
\section{Conclusion}

In this work, we introduced LTCF-Net, an innovative low-light image enhancement model that combines dual-channel color spaces with Transformer and Fourier Transform techniques to address the limitations of current enhancement methods. By effectively separating and processing illumination and color information, our model simplifies the enhancement process, allowing for end-to-end, single-stage training that improves operational efficiency and reduces the likelihood of artifacts such as noise and color distortion. Experimental results show that LTCF-Net achieves superior performance compared to existing state-of-the-art methods, as demonstrated on several benchmarks where it showed notable improvements in both quantitative metrics and qualitative assessments. Through extensive evaluations, LTCF-Net not only sets a new standard in low-light image enhancement but also suggests a promising direction for future research in integrating color space knowledge with deep learning architectures to further advance the field of image processing.